\documentclass{opt2026} % Include author names
\title{Subspace Levenberg–Marquardt Algorithms in Training Neural Networks}

\optauthor{%
\Name{M. Duc Hoang} \Email{mdhoang@ucdavis.edu}\\
\addr Department of Mathematics, University of California, Davis, Davis, CA, USA}

\usepackage{float}
\usepackage{indentfirst}
\begin{document}

\maketitle

\begin{abstract}%
The Levenberg-Marquardt (LM) algorithm is a well-known second-order method for rapid convergence and strong robustness when training small- to medium-sized neural networks (NNs). However, its computational and memory costs increase significantly as the number of parameters in an NN grows. To address this limitation, subspace methods have been proposed, such as the Krylov subspace LM (KSLM) and the hybrid subspace LM (HSLM), making second-order algorithms more efficient. In this work, we evaluate the subspace Levenberg-Marquardt algorithms for regression and classification tasks in neural networks. We compare the performance of subspace LM variants with the classical LM method, as well as other popular first-order algorithms, such as stochastic gradient descent (SGD) and Adam. 
\end{abstract}

% \begin{keywords}%
%  Levenberg-Marquardt, nonlinear least-square, subspace method, large-scale problem
% \end{keywords}
\section{Introduction}
Feedforward multilayer perceptrons are widely used for supervised learning tasks such as regression and classification. Their performance depends in part on whether the network has sufficient capacity to represent the underlying structure of the problem. Classical results show that feedforward networks with sufficiently many hidden units can approximate broad classes of nonlinear functions ~\cite{cybenko1989,hornik1989}. Approximation theory further shows that the network complexity required to achieve a prescribed accuracy can depend on properties of the target function, including its dimensionality and smoothness ~\cite{mhaskar1996,yarotsky2017}. Thus, more complex learning problems may require greater network capacity. In regression, this additional capacity allows the representation of richer nonlinear mappings, while in classification, increasing the number of parameters and layers can increase the capacity of the classifier ~\cite{bartlett2019}. However, larger networks also increase the number of trainable parameters and consequently make the optimization problem more computationally demanding.

First-order optimization methods, such as stochastic gradient descent (SGD) and Adam ~\cite{kingma2015}, are widely used for training neural networks due to their low memory requirements and relatively inexpensive iterations. However, their performance can depend strongly on hyperparameter choices, including the learning rate and momentum parameters. In addition, they may require a large number of iterations to achieve high accuracy, particularly when the optimization landscape is poorly conditioned. In contrast, second-order methods incorporate curvature information to determine more effective search directions and can achieve faster local convergence and greater robustness to problem scaling. For nonlinear least-squares training, the Levenberg–Marquardt (LM) method offers a practical alternative by using the Gauss–Newton approximation with adaptive damping and is particularly effective for small- to medium-sized networks ~\cite{hagan1994}.
However, the computational and memory costs associated with forming and solving the full LM system make it impractical as network size increases \cite{wilamowski2010}.

One approach for reducing these costs is to restrict the LM step to a low-dimensional subspace rather than solving the full-dimensional problem. Subspace LM methods retain important curvature information while reducing the size of the linear system that must be constructed and solved at each iteration. Subspace LM methods include Krylov subspace LM (KSLM) ~\cite{MizutaniDemmel2001}, which constructs a search space using Krylov directions. More recently, Hoang et al. introduced adaptive hybrid subspace LM (HSLM) ~\cite{HoangLewis2026}, which combines multiple informative directions to improve the quality of the reduced-space approximation. These approaches provide a potential compromise between the inexpensive iterations of first-order methods and the rapid convergence of full LM.

In this work, we investigate the use of subspace LM algorithms for both regression and classification tasks in neural networks. We compare classical LM with its subspace variants, KSLM and HSLM, and evaluate their performance against widely used first-order optimizers such as SGD and Adam. Our study focuses on the trade-offs among convergence behavior, computational cost, and predictive performance to assess whether subspace formulations can extend the practical advantages of LM to larger neural-network optimization problems.

\section{Levenberg–Marquardt and Subspace Methods for Neural Networks}
We consider the standard Levenberg-Marquardt algorithm and its subspace KSLM and HSLM variants.

\subsection{Levenberg--Marquardt Algorithm}
Nonlinear least-squares problems are commonly formulated as
\[
 \min_{x\in \mathbf{R}^n} \mathbf{F}(x) := \min_{x\in \mathbf{R}^n} \dfrac{1}{2} \|\mathbf{r}(x)\|^2,
\]
where $\mathbf{r}:\mathbb{R}^n\to\mathbb{R}^m$ is the residual vector. Let $\mathbf{J} \in \mathbb{R}^{m \times n}$ denote the Jacobian, then the Gauss--Newton method \cite{DennisSchnabel1996} solves nonlinear least-squares problems at iteration $k$ by linearizing the residual as $\mathbf{r}(x_k+s_k)\approx\mathbf{r}(x_k)+\mathbf{J}_ks_k$, resulting in a linear least-squares subproblem
\[
\left(\mathbf{J}_k^\top \mathbf{J}_k\right)s_k = -\mathbf{J}_k^\top\mathbf{r}(x_k)
\]
where $s_k$ is the update direction and $\mathbf{J}^\top \mathbf{J} \approx \mathbf{H} $ is the Gauss-Newton approximation to the Hessian. The Levenberg--Marquardt method can be viewed as a damped Gauss--Newton method for nonlinear least-squares problems \cite{Levenberg1944,Marquardt1963,More1978}. The LM step \(s_k\) is obtained by solving
\begin{equation}
\left(\mathbf{J}_k^\top \mathbf{J}_k + \mu_k \mathbf{I}_n\right)s_k = -\mathbf{g}_k ,
\label{eq:LM_classical}
\end{equation}
where $\mathbf{g}_k = \mathbf{J}_k^\top\mathbf{r}(x_k) $ is the gradient and $\mu_k>0$ is a damping parameter adjusted at each iteration. When $\mu_k$ is small, the method behaves similarly to Gauss--Newton, whereas for larger values the step approaches a scaled gradient-descent direction. This damping improves robustness when the Jacobian is ill-conditioned or the current iterate is far from a solution. See Appendix A for a detailed damping update strategy.

\subsection{Krylov-Subspace Levenberg--Marquardt Algorithm}
For large-scale nonlinear least-squares problems, solving the full LM system may become computationally expensive because forming and factorizing \(\mathbf{J}_k^\top \mathbf{J}_k + \mu_k \mathbf{I}_n\) can require substantial memory and work. In a Krylov-subspace LM approach, the step direction \(s_k\) is restricted to a subspace of the form
\[
\mathcal{K}_p(\mathbf{B}_k,\mathbf{g}_k)=\mathrm{span}\{\mathbf{g}_k,\mathbf{B}_k \mathbf{g}_k,\mathbf{B}_k^2 \mathbf{g}_k,\dots,\mathbf{B}_k^{p-1}\mathbf{g}_k\}, \qquad
\mathbf{B}_k = \mathbf{J}_k^\top \mathbf{J}_k + \mu_k \mathbf{I}_n.
\]
Let \(\mathbf{V}_k\in\mathbb{R}^{n\times p}\) be a matrix whose columns form a basis of \(\mathcal{K}_p(\mathbf{B}_k,\mathbf{g}_k)\). The step is then written as $s_k = \mathbf{V}_k y_k$, and \(y_k\in\mathbb{R}^p\) is obtained from the reduced system
\[
\left(\mathbf{V}_k^\top \mathbf{J}_k^\top \mathbf{J}_k\mathbf{V}_k+\mu_k \mathbf{I}_p\right)y_k=-\mathbf{V}_k^\top \mathbf{g}_k.
\]

Thus, instead of solving the full \(n\)-dimensional LM system, one solves a much smaller projected problem in the Krylov subspace. This approach can substantially reduce memory and computational requirements while retaining the LM damping mechanism, making it attractive for large neural-network problems where the full LM system is too expensive to solve directly \cite{MizutaniDemmel2001}. An additional advantage is that the Krylov subspace is invariant under the scalar shift introduced by the damping parameter \cite{Lin2016}. For fixed $\mathbf{J}_k$ and $\mathbf{g}_k$,
\begin{equation}
\mathcal{K}_p
\left(
\mathbf{J}_k^\top \mathbf{J}_k + \mu_k \mathbf{I}_n,
\mathbf{g}_k
\right)
=
\mathcal{K}_p
\left(
\mathbf{J}_k^\top \mathbf{J}_k,
\mathbf{g}_k
\right).
\end{equation}
Therefore, if several damping parameters are adjusted at the same iteration, the Krylov basis does not need to be recomputed. Thus, the basis can be reused while the $\mu_k$ is adjusted, further reducing the cost of repeated LM step computations.

\subsection{Hybrid Subspace Levenberg-Marquardt Algorithm}
HSLM adaptively constructs a low-dimensional subspace from multiple sources of gradient and curvature information. Specifically, this subspace combines multiple sources of basis information, including 
\begin{itemize}
\item the normalized gradient, which provides first-order descent information; 
\item recent optimization steps, which encode local optimization history; 
\item Krylov/Lanczos vectors, which approximate dominant curvature directions; and 
\item randomized Gauss-Newton-vector probes $(\mathbf{J}_k^\top \mathbf{J}_k) w$ where $w\sim \mathcal{N}(\mathbf{0},\mathbf{I}_n)$, which broaden spectral coverage and improve robustness when the local curvature is noisy or heterogeneous. 
\end{itemize}

Since the candidate pool combines complementary information, the resulting subspace may not fully capture the current descent direction. To assess its quality, a deterministic adequacy criterion is based on the fraction of gradient energy captured by the subspace
\begin{equation}\label{eq:eta_sub}
\eta^{\mathrm{sub}}_k := \frac{\|\mathbf{V}_k^\top \mathbf{g}_k\|^2}{\|\mathbf{g}_k\|^2}\in[0,1].
\end{equation}
The quantity $\eta^{\mathrm{sub}}_k \in[0,1]$ quantifies how much descent-relevant gradient information is captured by the current subspace, with larger values indicating greater adequacy.  If $\eta^{\mathrm{sub}}_k$ drops below a prescribed threshold $\eta_{\min}\in(0,1]$, additional candidate directions are added and the adequacy test is repeated until the threshold is satisfied. Once the adequacy criterion is satisfied, the reduced basis is fixed and the Levenberg--Marquardt step is computed entirely within this subspace, reducing the original $n$-dimensional problem to a much smaller $p$-dimensional system. 

Let $\mathbf{J}_{k,p}=\mathbf{J}_k\mathbf{V}_k$ be the reduced Jacobian. Let the singular value decomposition of $\mathbf{J}_{k,p}$ be
\[
\mathbf{J}_{k,p} = \mathbf{U}_k\boldsymbol{\Sigma}_{k,p} \mathbf{Z}_k^\top,
\qquad
\boldsymbol{\Sigma}_{k,p}=\mathrm{diag}(\sigma_{k,1},\dots,\sigma_{k,p}).
\]
Writing the search direction as $s_k=\mathbf{V}_k\mathbf{Z}_k y_k$, the resulting LM system can then be solved efficiently in the singular-vector coordinates as ~\cite{HoangLewis2026}
\[
\left(\boldsymbol{\Sigma}_{k,p}^{2}+\mu_k\mathbf{I}_p\right)y_k
=
-\boldsymbol{\Sigma}_{k,p}\mathbf{U}_k^\top\mathbf{r}_k.
\]
This reduced formulation inherits the isotropic damping of the standard LM algorithm. However, because the singular values of the reduced Jacobian can vary substantially, treating all directions equally may lead to unnecessary damping in some directions and insufficient damping in others. To better accommodate ill-conditioning and variations in local curvature, the HSLM replace the isotropic damping matrix $\mathbf{I}_p$ with a diagonal, curvature-dependent damping matrix $\mathbf{D}_{k,p}$. The resulting reduced system is
\[
\mathbf{B}_{k,p}\mathbf{y}_k
=
-\boldsymbol{\Sigma}_{k,p}\mathbf{U}_k^\top\mathbf{r}_k,
\]
where
\[
\mathbf{B}_{k,p}
=
\boldsymbol{\Sigma}_{k,p}^{2}
+
\mu_k\mathbf{D}_{k,p},
\qquad
\mathbf{D}_{k,p}
=
\operatorname{diag}(d_{k,1},\ldots,d_{k,p}),
\qquad
d_{k,i}
=
\max(\sigma_{k,i}^{2},\delta).
\]
The threshold $\delta>0$ keeps the damping coefficient strictly positive, while singular value dependence provides stronger damping in directions with greater curvature.

Unlike the classical LM algorithm and KSLM, HSLM uses Armijo backtracking for step acceptance, with the damping parameter adjusted according to whether the step is accepted or rejected. The HSLM method generalizes subspace optimization by combining multiple sources of information within an adaptive low-dimensional space, while the deterministic adequacy strategy ensures that sufficient descent-relevant information is retained. This allows the method to reduce computational cost while preserving the main theoretical advantages of LM-type methods ~\cite{HoangLewis2026}.

\section{Evaluation and Comparison}
We evaluate the optimization methods on two neural-network problems: a nonlinear regression problem and the 13-bit parity classification problem. In both experiments, we compare SGD, Adam, the classical  LM method, KSLM, and HSLM. Each experiment is repeated over 100 independent trials using common initialization across methods. See Appendix A for more implementation details. 

\subsection{Damped Oscillation Regression Problem}
All five methods successfully approximate the damped-oscillation function, $f(x)=2e^{-x^2}\cos(2\pi x)$ (Figure~1A), but differ significantly in convergence speed and computational cost (Figure~1B and Table~1). LM and KSLM converge in relatively few iterations, averaging $22.2 \pm 6.8$, but require higher execution times of $83.0 \pm 26.2$ s and $132.8 \pm 40.5$ s, respectively. While LM computes updates in the full parameter space, KSLM and HSLM restrict the step to lower-dimensional subspaces. KSLM retains fast LM-like convergence but incurs additional cost from constructing and expanding the Krylov subspace. HSLM achieves a better balance, requiring only $19.3 \pm 5.2$ iterations and $25.9 \pm 7.4$ s on average. In contrast, SGD and Adam require substantially more epochs before termination, averaging $155.0 \pm 55.1$ and $102.1 \pm 44.3$, respectively.
\begin{figure}[H]
\centering
\includegraphics[width=0.8\linewidth]{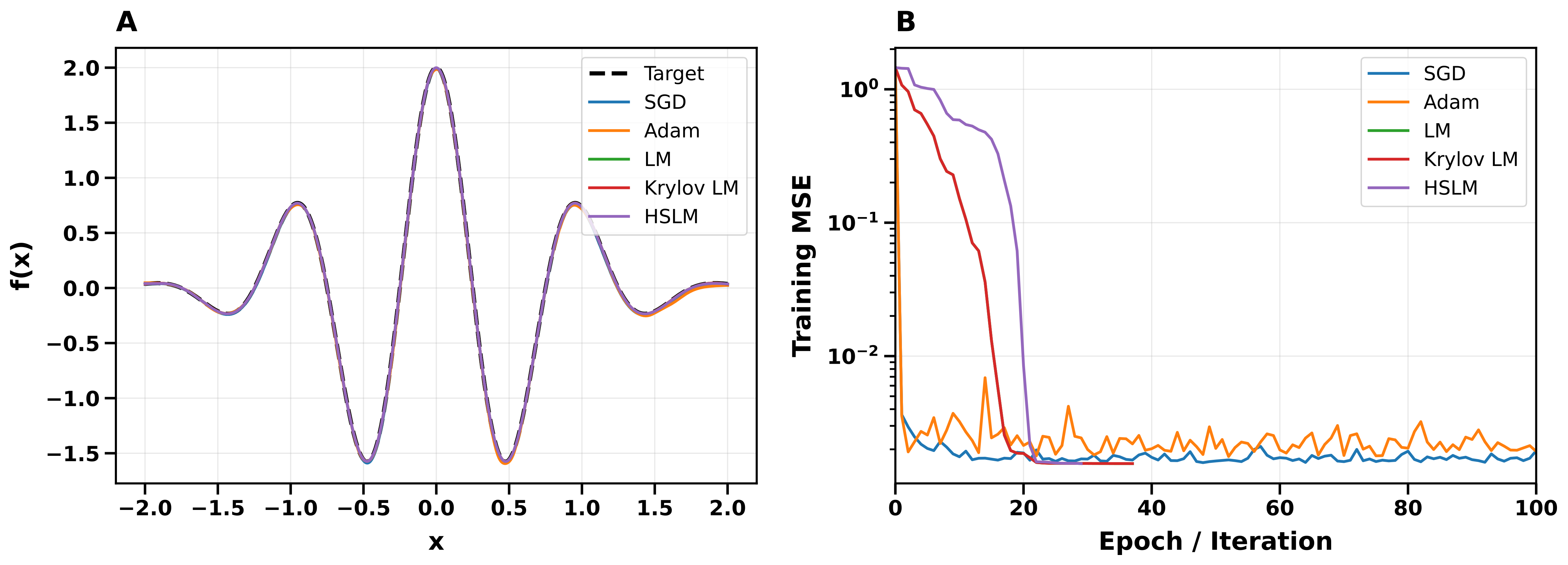}
\vspace{-10pt}
\caption{\small Comparison of optimization methods on Damped Oscillation regression problem. A. Approximation of the target function obtained using SGD, Adam, LM, Krylov LM, and HSLM. B. Training mean squared error (MSE) over 100 epochs/iterations. See Appendix B1 for the complete training performance.}
\label{fig:subim1}
\end{figure} 
\vspace{-25pt}
\begin{table}[H]
\centering
\small
\setlength{\tabcolsep}{3pt}
\renewcommand{\arraystretch}{1.05}

\begin{tabular}{|l|c|c|c|c|c|}
\hline
& LM & KSLM & HSLM & SGD & Adam \\
\hline
Execution Time (s)
& 83.0 $\pm$ 26.2
& 132.8 $\pm$ 40.5
& 25.9 $\pm$ 7.4
& 30.9 $\pm$ 10.9
& 24.3 $\pm$ 10.6 \\
\hline
Iteration/Epoch
& 22.2 $\pm$ 6.8
& 22.2 $\pm$ 6.8
& 19.3 $\pm$ 5.2
& 155.0 $\pm$ 55.1
& 102.1 $\pm$ 44.3 \\
\hline
Training MSE $\leq \sigma_{\mathrm{noise}}^2$ (\%)
& 100
& 100
& 100
& 41
& 0 \\
\hline
\end{tabular}
\vspace{-5pt}
\caption{\small Summary of training performance for neural-network algorithms across 100 trials (mean $\pm$ SD).}
\label{tab:nn_training_1}
\end{table}
\vspace{-10pt}
Moreover, all three LM-based methods achieved a 100\% success rate in reaching the prescribed noise-variance threshold, $\sigma_{\mathrm{noise}}^2 \approx 0.0016$, compared with 41\% for SGD and 0\% for Adam. By exploiting curvature information and adaptive damping, LM-based methods make stable, well-scaled updates near the minimum and more reliably approach the noise floor. In contrast, first-order methods may progress more slowly near the minimum. Among the LM-based methods, HSLM retains this fast and reliable convergence while substantially reducing the computational cost of classical LM and KSLM.
\subsection{13-Parity Classification Problem}
For the 13-bit parity problem, all methods achieved high training and validation accuracy but differed substantially in convergence behavior (Figure~2 and Table~2). The LM-based methods reduced training and validation MSE faster than SGD and Adam (Figure~2A--B). HSLM required the fewest outer iterations among the LM-based methods, $37.0 \pm 27.2$ iterations on average, while SGD and Adam required substantially more epochs, $1061.7 \pm 530.1$ and $333.3 \pm 458.5$.

\begin{figure}[H]
\centering\includegraphics[width=0.728\linewidth]{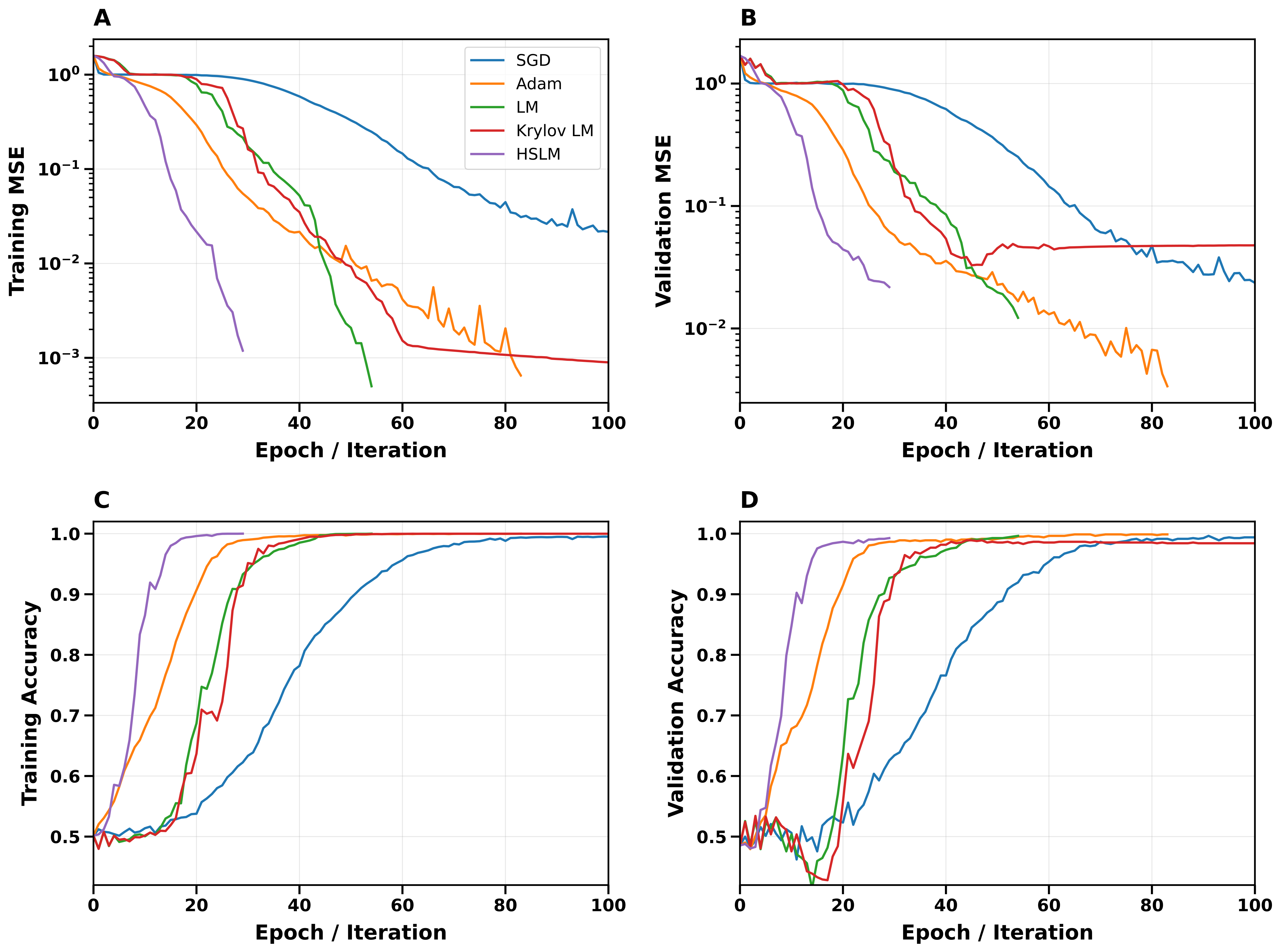}
\vspace{-10pt}
\caption{\small Training and validation performance of the optimization methods on the 13-parity classification problem over 100 epochs/iterations. A. Training MSE. B. Validation MSE. C. Training accuracy. D. Validation accuracy. See Appendix B2 for the complete training performance}
\label{fig:subim2}
\end{figure} 
\vspace{-25pt}

\begin{table}[H]
\centering
\small
\setlength{\tabcolsep}{3pt}
\renewcommand{\arraystretch}{1.05}

\begin{tabular}{|l|c|c|c|c|c|}
\hline
& LM & KSLM & HSLM & SGD & Adam \\
\hline
Execution Time (s)
& 5.8 $\pm$ 4.1
& 6.3 $\pm$ 4.0
& 3.7 $\pm$ 3.8
& 13.2 $\pm$ 6.5
& 5.7 $\pm$ 7.8 \\
\hline
Iteration/Epoch
& 67.8 $\pm$ 47.8
& 71.1 $\pm$ 46.3
& 37.0 $\pm$ 27.2
& 1061.7 $\pm$ 530.1
& 333.3 $\pm$ 458.5 \\
\hline
Validation accuracy (\%)
& 97.7 $\pm$ 7.7
& 97.9 $\pm$ 5.1
& 99.6 $\pm$ 0.3
& 99.7 $\pm$ 0.3
& 99.8 $\pm$ 0.1 \\
\hline
Convergence rate (\%)
& 92
& 89
& 100
& 92
& 100 \\
\hline
\end{tabular}
\vspace{-5pt}
\caption{\small Comparison of the training performance of the optimization methods over 100 trials (mean $\pm$ SD). Convergence is defined as achieving a training MSE below $0.01$ and a training accuracy above $0.99$.}
\label{tab:nn_training_2}
\end{table}
\vspace{-10pt}
HSLM and Adam achieved a 100\% convergence rate, compared with 92\%, 89\%, and 92\% for LM, KSLM, and SGD, respectively. Adam obtained the highest validation accuracy ($99.8 \pm 0.1\%$), while HSLM achieved a comparable $99.6 \pm 0.3\%$ with far fewer iterations. HSLM also remains effective with limited training data (see Appendix C), providing a strong balance of convergence speed, computational cost, and predictive accuracy.
\section{Conclusion and Future Work}
The numerical results show that the HSLM method achieves a favorable balance of convergence speed, computational efficiency, and solution accuracy for both regression and classification problems, suggesting that reduced-subspace approaches can preserve the fast convergence of LM-type methods while substantially reducing computational cost for neural-network training. However, larger datasets may increase the cost of forming and solving the associated subproblems, limiting scalability. Future work may explore broader classes of subspace methods that incorporate mini-batch and matrix-free implementations, and investigate how different choices of subspace basis affect computational efficiency and convergence behavior.

\bibliography{sample}
\newpage
%\clearpage
\appendix
\section{Experimental Setup}
We evaluate five optimization methods---SGD, Adam, classical LM, Krylov-subspace LM, and HSLM---on nonlinear regression and 13-bit parity classification. Each experiment is repeated over 100 independent trials, with parameters initialized as $\theta_0 \sim \mathcal{U}[-1,1]$. Within each trial, all methods use the same initial parameter vector to ensure a fair comparison. All experiments were implemented in Python using NumPy within Jupyter Notebook and executed on a Dell 14 Plus laptop equipped with an Intel Core Ultra 9 288V processor and 32 GB of RAM.
\subsection{Regression problem}
For the regression task, we approximate the function
\[
    f(x) = 2e^{-x^2}\cos(2\pi x),
    \qquad x\in[-2,2].
\]

We generate 40,000 randomly sampled input points and corrupt the corresponding function values with additive Gaussian noise with a standard deviation equal to 5\% of the standard deviation of the target value, $\sigma_{y}$,
\[
    y_i = f(x_i) + \sigma_{\mathrm{noise}}\epsilon_i,
    \qquad
    \epsilon_i \sim \mathcal{N}(0,1),  
     \qquad
    \sigma_{\mathrm{noise}}
    = 0.05 \sigma_{y}
\]

The regression model is a fully connected multilayer perceptron with architecture $1\!-\!70\!-\!40\!-\!1$. The two hidden layers use the hyperbolic tangent ($\tanh$) activation function, while the output layer is linear. Mean squared error (MSE) is used as the loss function. A single realization of the input data and Gaussian noise is generated and then kept fixed across all optimizers and all 100 trials.

\subsection{13-bit parity problem}
For the classification experiment, we consider the complete 13-bit parity problem, consisting of all $2^{13}=8192$ binary input patterns. Both the inputs and target labels are represented using $\{-1,1\}$. We use a fully connected MLP with architecture $13\!-\!25\!-\!10\!-\!1$, with $\tanh$ activation functions in both hidden layers and the output layer. The network is trained using MSE as the loss function. We split the full parity dataset using a stratified 90/10 training--validation split, fixed across optimization methods and all 100 trials.

\subsection{First-order methods}
For SGD, we use a learning rate of $0.01$, momentum of $0.9$, and a mini-batch size of 64. For Adam, we use a learning rate of $0.003$ and a mini-batch size of 64, with the standard momentum parameters $\beta_1=0.9$ and $\beta_2=0.999$. Both SGD and Adam are allowed a maximum of 1500 epochs.

\subsection{LM-based methods}
For classical LM, KSLM, and HSLM, the maximum number of outer iterations is set to 150. The initial damping parameter is $ \mu_0 = 10$. In this work, the damping parameter follows the delayed-gratification strategy \cite{Transtrum2011}:
\[
\mu_{k+1} =
\begin{cases}
 \mu_k/\mu_{\mathrm{down}}, & \text{if successful} \\[4pt]
 \mu_k \cdot \mu_{\mathrm{up}}, & \text{otherwise}\
\end{cases}
\]

This allows the LM method to move more quickly toward the Gauss--Newton regime when the local model is reliable, while shifting toward the gradient-descent regime when a step is unsuccessful. Following a successful step, the damping parameter is reduced according to $\mu_{k+1} = \mu_k/2$, whereas following an unsuccessful step it is increased according to $\mu_{k+1} = 5\mu_k$. For Krylov LM, the maximum Krylov-subspace dimension is restricted to 5\% of the full parameter dimension, and the Lanczos tolerance is set to $10^{-3}$. For HSLM, the initial random-probe subspace dimension is set to 1\% of the full parameter dimension. The subspace adequacy threshold is set to $\eta_{\min}=0.99$. When the current subspace is inadequate, it is expanded using Lanczos directions corresponding to 2\% of the parameter dimension and additional random Hessian-probe directions corresponding to 1\% of the parameter dimension. The total subspace dimension is restricted to at most 10\% of the full parameter dimension. See ~\cite{HoangLewis2026} for a more detailed implementation.

\subsection{Stopping criteria}
For the noisy regression problem, the target training error is defined by the variance of
noise, $ \mathrm{MSE}\leq \sigma_{\mathrm{noise}}^2$. For SGD and Adam, optimization is also terminated if the relative improvement in MSE remains below $10^{-5}$ for 50 consecutive epochs or if the maximum epoch budget is reached. For the LM-based methods, optimization may additionally terminate when $\left\|J^{T}r\right\|_{\infty} \leq 10^{-8}$, or when the step satisfies $\left\|s_k\right\|_{\infty} \leq 10^{-10} \left(1+\left\|\theta_k\right\|_{\infty}\right)$, or when the maximum iteration budget is reached.

For the 13-bit parity problem, the training stopping criterion requires $\mathrm{MSE} \leq 10^{-2}$ together with above 99\% training accuracy. We record the training and validation MSE, classification accuracy, number of iterations or epochs, and execution time for each optimizer over the 100 trials.

\section{Detailed Training Performance}
\subsection{Damped Oscillation Regression Problem}
\begin{figure}[H]
\centering\includegraphics[width=0.7\linewidth]{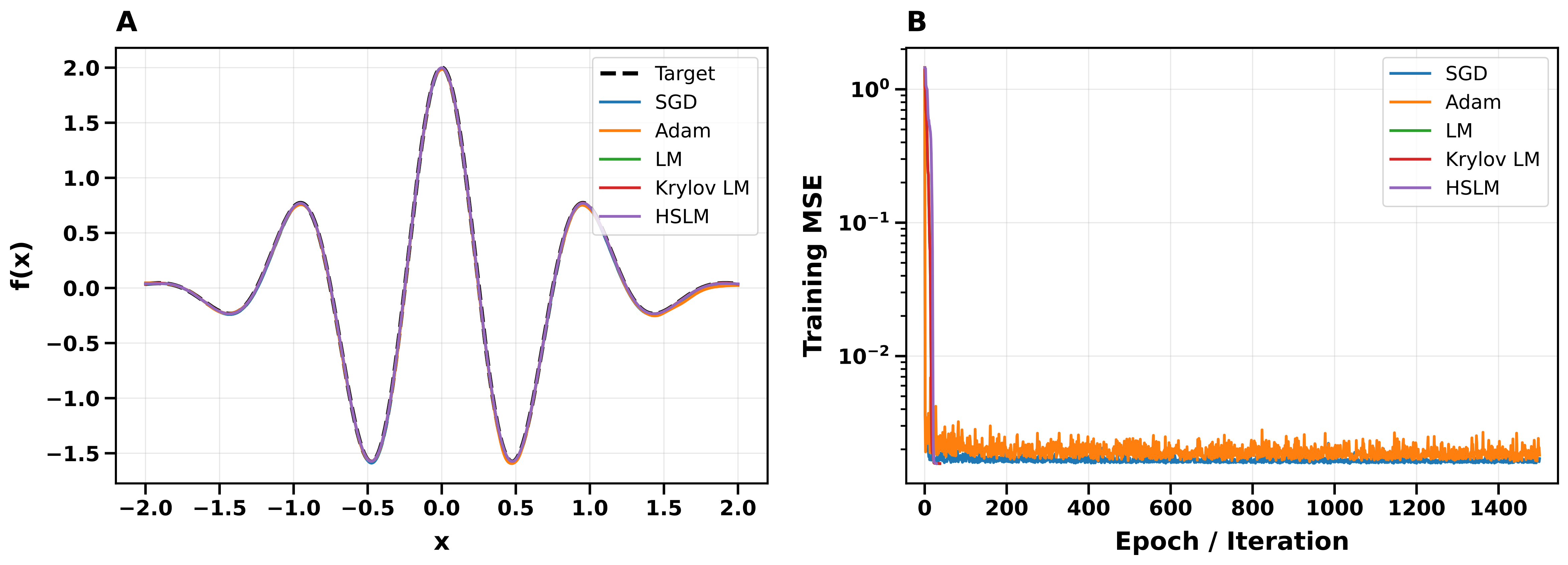}
\caption{Comparison of optimization methods on Damped Oscillation regression problem. A. Approximation of the target function obtained using SGD, Adam, LM, Krylov LM, and HSLM. B. Training mean squared error (MSE) vs epoch/iteration for optimization methods.}
\label{fig:subS1}
\end{figure} 

\subsection{13-Parity Classification Problem}
\begin{figure}[H]
\centering\includegraphics[width=0.7\linewidth]{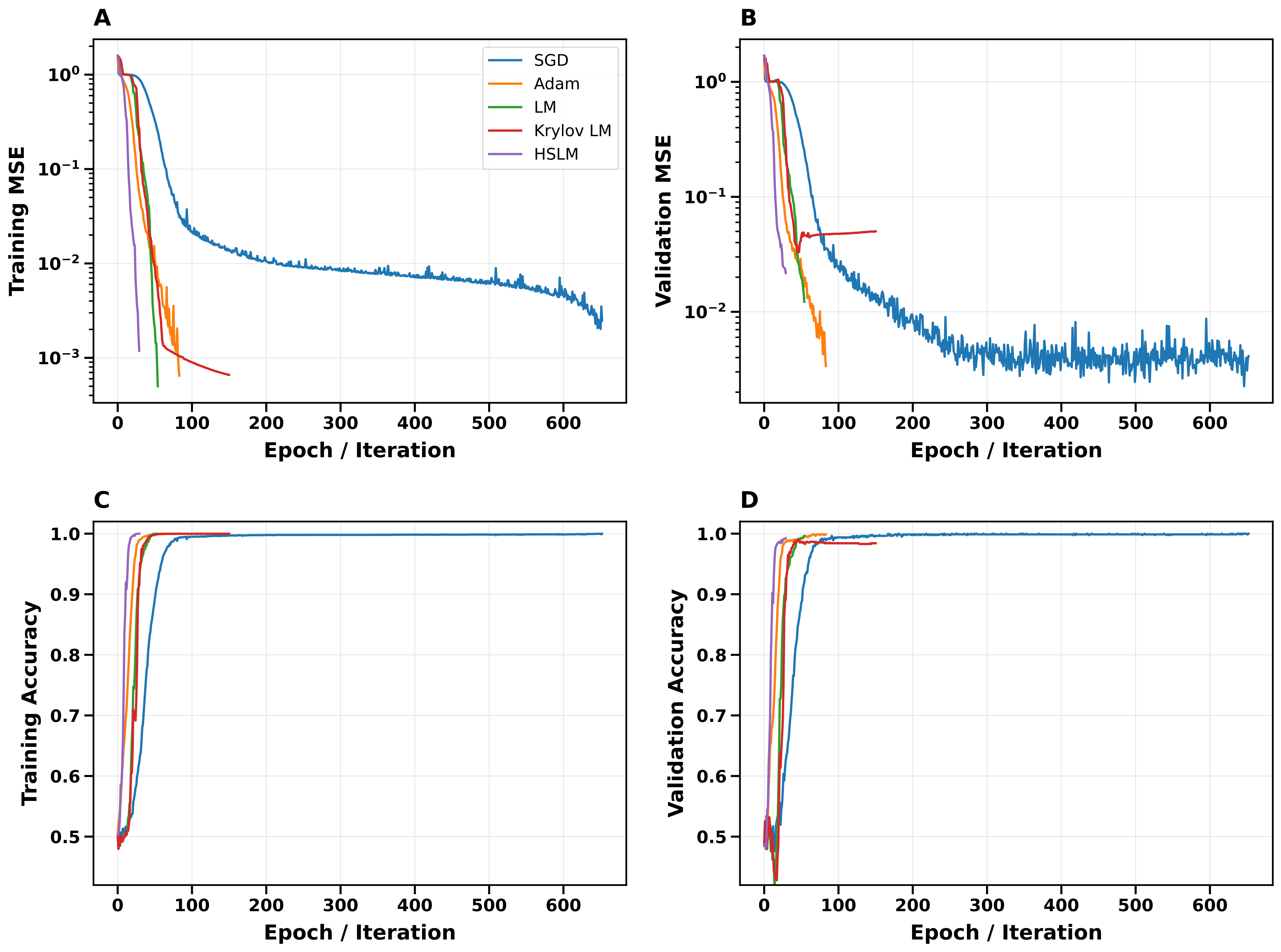}
\caption{Training and validation performance of the optimization methods on the 13-parity classification problem. 
A. Training MSE. B. Validation MSE. C. Training accuracy. D. Validation accuracy.}
\label{fig:subS2}
\end{figure} 

\section{Further Experiments}
To evaluate optimizer performance under limited-data conditions, we reduced the training/validation split to 30/70 and reran the algorithms on the 13-bit parity problem. The results show that the HSLM method maintains a stronger balance between convergence speed, computational efficiency, and predictive accuracy compared to other methods.
\begin{figure}[H]
\centering\includegraphics[width=0.7\linewidth]{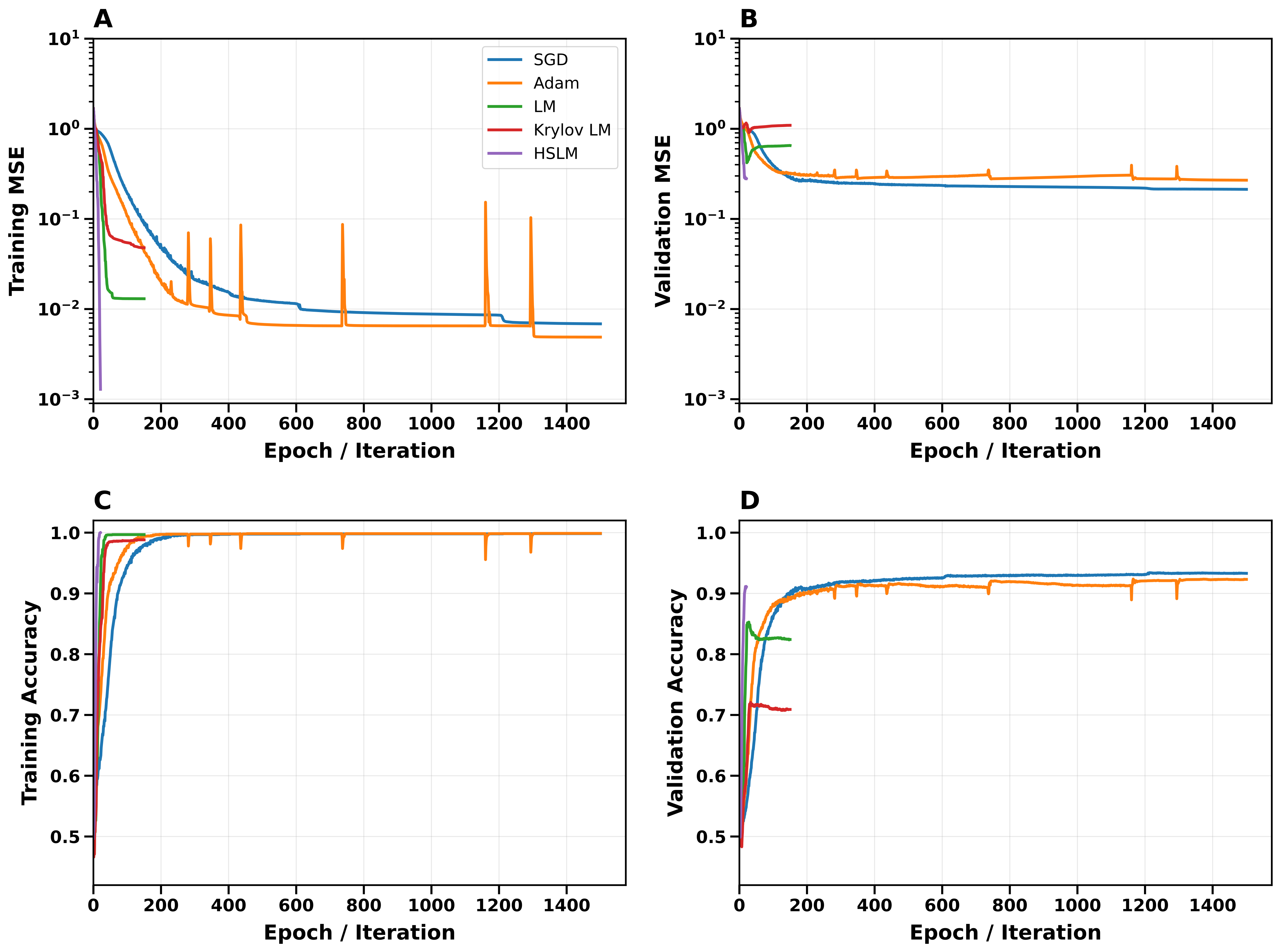}
\caption{Training and validation performance of the optimization methods on the 13-parity classification problem. 
A. Training MSE. B. Validation MSE. C. Training accuracy. D. Validation accuracy.}
\label{fig:subS3}
\end{figure} 

\end{document}